\documentclass[sigconf]{acmart}

\AtBeginDocument{%
  \providecommand\BibTeX{{%
    \normalfont B\kern-0.5em{\scshape i\kern-0.25em b}\kern-0.8em\TeX}}}

\usepackage{makecell}
\usepackage{multirow}
\usepackage{color}
\usepackage{enumitem}
\usepackage{url}
\setlist[itemize]{leftmargin=*}
\setcopyright{acmcopyright}

\copyrightyear{2021}
\acmYear{2021}
\setcopyright{acmcopyright}\acmConference[ICMR '21]{Proceedings of the 2021 International Conference on Multimedia Retrieval}{August 21--24, 2021}{Taipei, Taiwan}
\acmBooktitle{Proceedings of the 2021 International Conference on Multimedia Retrieval (ICMR '21), August 21--24, 2021, Taipei, Taiwan}
\acmPrice{15.00}
\acmDOI{10.1145/3460426.3463598}
\acmISBN{978-1-4503-8463-6/21/08}

\begin{document}
\fancyhead{}

\title{Reproducibility Companion Paper: Knowledge Enhanced Neural Fashion Trend Forecasting}

 \author{Yunshan Ma}
 \affiliation{
     \institution{National University of Singapore}
 }
 \email{yunshan.ma@u.nus.edu}
 
 \author{Yujuan Ding}
 \affiliation{
     \institution{Shenzhen University}
 }
 \email{dingyujuan385@gmail.com}
 
 \author{Xun Yang}
 \affiliation{
     \institution{National University of Singapore}
 }
 \email{xunyang@nus.edu.sg}

 \author{Lizi Liao}
 \affiliation{
     \institution{National University of Singapore}
 }
 \email{liaolizi.llz@gmail.com}

 \author{Wai Keung Wong}
 \affiliation{
    \institution{The Hong Kong Polytechnic University}
 }
 \email{calvin.wong@polyu.edu.hk}

 \author{Tat-Seng Chua}
 \affiliation{
     \institution{National University of Singapore}
 }
 \email{dcscts@nus.edu.sg}
 
 \author{Jinyoung Moon}
 \affiliation{
     \institution{ETRI, South Korea}
     \institution{UST, South Korea}
 }
 \email{jymoon@etri.re.kr}
 
 \author{Hong-Han Shuai}
 \affiliation{
     \institution{National Yang Ming Chiao Tung University, Taiwan}
 }
 \email{hhshuai@nctu.edu.tw}

\begin{abstract}
This companion paper supports the replication of the fashion trend forecasting experiments with the KERN (Knowledge Enhanced Recurrent Network) method that we presented in the ICMR 2020. We provide an artifact that allows the replication of the experiments using a Python implementation. The artifact is easy to deploy with simple installation, training and evaluation. We reproduce the experiments conducted in the original paper and obtain similar performance as previously reported. The replication results of the experiments support the main claims in the original paper.

\end{abstract}



\maketitle
\section{Introduction}
In the original paper~\cite{ma2020knowledge}, we presented the KERN model for addressing the fashion trend forecasting problem. The problem aims to take a period of historical fashion trend records as input, and predict a period of following trends in the future. It is formulated as a multi-horizon time-series forecasting task. Our KERN method utilizes a deep learning-based framework, \textit{i.e.}, LSTM encoder-decoder, and further incorporates both internal and external knowledge to enhance the forecasting performance. In this companion paper, we present a replication artifact that provides a complete re-implementation of the model as well as the required experimental facilities used in the original paper.

\begin{figure}[!tp]
	\centering
	\includegraphics[scale = 0.42]{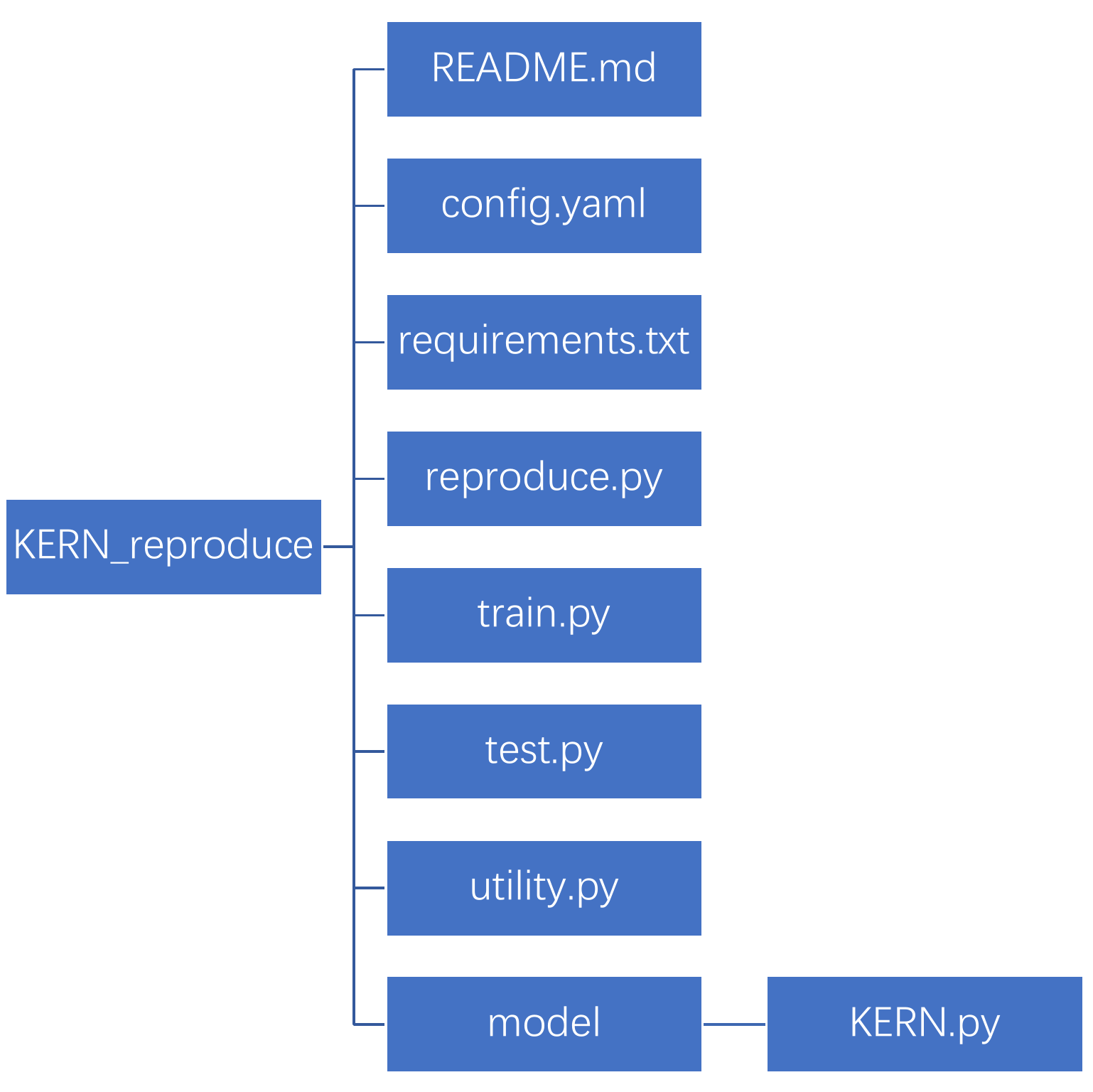}
	\caption{The structure of KERN\_reproduce artifact.
	}
	\label{fig1}
\end{figure}

\renewcommand\arraystretch{1.3}
\begin{table}
\centering
  \caption{Dataset setting}
  \label{tab:dataset}
    \begin{tabular}{ccccc}
    \toprule
    \multirow{2}*{{Dataset}} & Input length  & Output length  & \#training & \#test \\
    ~ & (year)  & (year) & sample & sample \\
    \hline
    GeoStyle &1 &0.5 &80,960 &2,024 \\
    \hline
    \multirow{2}*{{FIT}} &2 &0.5 &422,784 &8,808 \\ 
    ~ &2 &1 &211,392 &8,808 \\
    \bottomrule
  \end{tabular}
\end{table}

\begin{table*}[!t]
\caption{Introduction of Parameters}
  \begin{tabular}{p{2.5cm}p{11cm}p{1.5cm}p{1cm}}
\toprule
\multirow{2}*{{Parameter}} &\multirow{2}*{{description}} &\multicolumn{2}{c}{Default value}\\
~ &~ &FIT &GeoStyle\\
\midrule
input\_len  & the length of input fashion trend sequence (for encoder) & 48 &52\\
output\_len   & the length of the fashion trend sequence to predict (for decoder) & 12/24 & 26\\

ext\_kg  & whether to apply the external knowledge & true & true \\
int\_kg  & whether to apply the internal knowledge & true & true \\
triplet\_lambda  & coefficient for the triplet loss (chosen from [0.0001,0.0002,0.001,0.002,0.01,0.02]) & 0.002 & 0.002 \\
sample\_range  & the range to sample similar and dissimilar sequence(choose from [50,100,500,1000])  & 500  & 500\\
feat\_size  & the embedding dimension for the element and user features & 10 & 10\\
rnn\_hidden\_size  & the dimension of the LSTM hidden state &50 &50\\
lr  & the initialized learning rate & 0.001  & 0.001\\
lr\_decay  & whether to decay the learning rate during the training & true & true \\
lr\_decay\_interval  & after how many epochs to decay the learning rate &10 &15 \\
lr\_decay\_gamma  & the learning rate decay rate & 0.1 & 0.1\\
epoch  & the number of total training epochs & 15 & 20 \\
batch\_size  & the batch size & 400 & 400\\

 \bottomrule
\end{tabular}
\label{tab:parameter}
\end{table*}

The artifact \textit{KERN\_reproduce} is available at
\url{https://github.com/mysbupt/KERN_reproduce}. 
The dataset is accessible at
\url{https://drive.google.com/file/d/1E_gPmh6lIHyXEx3lRDWWKvm_8op5B82t/view?usp=sharing}. We also pack and upload the code and dataset to the platform Zenodo, and the URL is \url{https://zenodo.org/record/4774766#.YKdfH6LnhhE}.
This artifact includes a detailed README file which specifies the environmental requirements, code structures and implement instructions. It also includes the code to train and evaluate the KERN model for fashion trend forecasting. The experimental settings are easy to configure by a unified configuration file in the artifact. The file structure of the artifacts is shown in Figure~\ref{fig1}.

\section{Dataset}
In the original paper, two datasets, \textit{i.e.} FIT~\cite{ma2020knowledge} and GeoStyle~\cite{mall2019geostyle}, are involved to evaluate the effectiveness of the presented method. In our replication, we use the exactly same datasets. Both of the two datasets consist of considerable number of specific fashion trend time-series. Each time-series denotes the evolving process of a specific type of fashion element among a certain group of people. In comparison, FIT has longer fashion trend sequences, includes more fine-grained fashion elements, and considers more user attributes than GeoStyle. The detailed settings and statistics of the two datasets in our experiments are shown in Table~\ref{tab:dataset}. The datasets which are directly applied in our experiments can be downloaded according to the instruction in the README file in the artifacts. The original GeoStyle dataset is also provided, along with the preprocessing code in the script.  

\section{Experimental Setup}
The code is purely written with Python and the required modules for using the artifact are listed in the file of \textit{requirements.txt}. In particular, we use the PyTorch (1.6.0 or higher version) framework to implement the model. In addition, we have tested the code on servers which have the OS of Ubuntu 16.04/18.04 and are equipped with Nvidia TiTan X/V/Xp/V100 GPUs. 

The KERN model is an end-to-end neural network model, which can be easily trained and evaluated by the proposed artifact. Here we describe the utility of each script in the artifact. 
\begin{itemize}
    \item \textbf{reproduce.py} The auxiliary script to reproduce all the experimental results reported in Table~\ref{tab:results1} and Table~\ref{tab:results2}.
    
    \item \textbf{train.py} The script to train the KERN method following the default settings defined in the configure file (./config.yaml). 
    
    \item \textbf{test.py} The script to generate the testing result for the saved models.
    
    \item \textbf{utility.py} The code for preparing data, including both dataset preprocessing and data loader. It generates training and testing samples which are fed into the model during training.
    
    \item \textbf{model/KERN.py} The code defines the model of KERN.
\end{itemize}

\renewcommand\arraystretch{1.3}
\begin{table*}[!htp]
  \caption{The original and the replication performance of KERN}
  \setlength{\tabcolsep}{6mm}{
  \begin{tabular}{ccc|cc|cc}
    \toprule
    Dataset &\multicolumn{2}{c|}{GeoStyle} &\multicolumn{4}{c}{FIT}  \\
    \midrule
    ~\multirow{2}*{\texttt{Method}} &\multicolumn{2}{c|}{Half year}  &\multicolumn{2}{c|}{Half year}  &\multicolumn{2}{c}{One year}\\
    ~ &MAE &MAPE  &MAE  &MAPE  &MAE  &MAPE   \\
    \midrule
    \textbf{Original} &0.0134 &14.24 &0.083 &30.02 &0.094 &33.45\\
    \textbf{Replication} &0.0129 &14.77 &0.082 &29.40 &0.093 &32.56 \\
    \bottomrule
  \end{tabular}}
  \label{tab:results1}
\end{table*}

\renewcommand\arraystretch{1.3}
\begin{table*}
  \caption{Comparison of the results of ablation study. (KERN-I: w/o internal knowledge, KERN-E:w/o external knowledge, KERN-IE:w/o both)}

  \label{tab:results2}
  \setlength{\tabcolsep}{4.2mm}{
  \begin{tabular}{ccccccc}
    \toprule
    Dataset &\multicolumn{2}{c}{GeoStyle} &\multicolumn{4}{c}{FIT}  \\
    \midrule
    Prediction &\multicolumn{2}{c}{Half year}  &\multicolumn{2}{c}{Half year}  &\multicolumn{2}{c}{One year}\\
    ~ &original &replication &original &replication &original &replication \\
    \midrule
    \texttt{KERN-IE} &0.0137 &0.0130  &0.0840 &0.0824 &0.0966 &0.0940  \\
    \texttt{KERN-E} &-   &- &0.0835 &0.0827 &0.0953 &0.0941 \\
    \texttt{KERN-I}&0.0137 &0.0130 &0.0831 &0.0824 &0.0942 &0.0940 \\
    \texttt{KERN} &0.0134 &0.0129 &0.0836 &0.0823 &0.0939 &0.0931 \\
    \bottomrule
  \end{tabular}}
\end{table*}

\vspace{-0.20in}
\section{Parameters}
We provide the \textit{config.yaml} file for users to adjust the hyper-parameters needed to reproduce the experiments. We list all the configured settings and hyper-parameters used in the experiments in Table~\ref{tab:parameter}, as well as the detailed descriptions and the default values. Note that some parameters are set differently for different experimental settings, such as \textit{triplet\_lambda} and \textit{epoch}. Parameters for model training, including \textit{lr}, \textit{epoch} and \textit{batch\_size}, are chosen based on empirical observations of the training process considering both convergence and stability, which we suggest users keep the default values for these settings in order to generate fair results. \textit{input\_len} and \textit{output\_len} are set differently for two datasets, resulting in three settings: \textit{FIT output\_len 12, FIT output\_len 24, GeoStyle output\_len 26}. \textit{feat\_size} and \textit{rnn\_hidden\_size} are model hyper-parameters, which are kept the same with the original paper. By adjusting \textit{ext\_kg} and \textit{int\_kg}, users can conduct various ablation studies to verify the effectiveness of introducing external knowledge and internal knowledge individually or collaboratively. We tune \textit{triplet\_lambda} (corresponds to $\lambda$ in equation (9) of the original paper) and \textit{sample\_range} for the KERN model and the ablated model with internal knowledge only to achieve the best performance. Specifically, the hyper-parameter \textit{sample\_range} is used to control the sampling of $p$ and $q$ in terms of a certain $k$, the smaller the \textit{sample\_range}, $p$ and $q$ will be more closer with each other.

\section{Experimental results}
We reproduce the experiments on two datasets for all settings applied in the original paper. The overall performance of the KERN model, both in the original paper and our replication, are reported in Table~\ref{tab:results1}. From the table, we can see that the replication results are not identical but quite similar with the original reported results. We carefully go through all the scripts used for the experiment, and find the reason might be the randomness introduced during the initialization of the deep learning model's parameters. 

We also reproduce the ablation experiments and report the original and replicated results in Table~\ref{tab:results2}. Specific settings for two hyper-parameters \textit{triplet\_lambda} and \textit{sample\_range} are: [KERN/KERN-I, GeoStyle]: [0.002, 50], [KERN-I, FIT-half year]:[0.0001, 500], [KERN, FIT-half year]:[0.001, 500], [KERN-I, FIT-one year]:[0.01, 1000], [KERN, FIT-one year]:[0.0002, 100]. Users can also find the detailed settings for various experiments in the \textit{config.yaml} file that we provide in the artifact. From the table, we can see that our replicated results are overall similar with the original reported ones. Although the results are not identical, most of our claims in the original paper still hold with the replicated results. For FIT, the external knowledge is helpful in improving the performance for both experimental settings. However, the internal knowledge is not consistent helpful in all settings. Based on our replicated results, when predicting the FIT for half year, adding the internal knowledge upon the external knowledge-involved model slightly degrades the final performance. All the results and conclusions obtained are consistent with those in the original paper. By the way, we find a presentation typo in the original paper, where the KERN-I result of GeoStyle in Table 2 should be 0.0137. In this paper, we fix this typo. 

\section{Reproducibility Efforts}
The last two co-authors of this paper are the reviewers of this companion paper. Through some corrections and improvements, we have successfully installed necessary libraries and execute source codes from scratch. Moreover, we have reproduced results that are quite close to ones from the original paper~\cite{ma2020knowledge}, We deeply appreciate the active efforts of the authors of the original paper to improve the reproducibility of this companion paper, which benefit the future research.

During the reviewing process, we focused on corrections and improvements on the user-friendly installation and execution that enable other researchers to obtain the reproduced results without struggling in the trial-and-error process. First, we provide additional explicit information on CUDA version and all necessary python libraries with their specific versions, which are described in \textit{requirements.txt}. Second, we add an auxiliary script of \textit{reproduce.py} that runs all the testing together to easily reproduce and check all the results. Finally, we separate testing codes from training codes (\textit{train.py}) to provide better development environment to other researchers for additional training and testing beyond reproducing the described results. We save the best model derived from the train.py and add an additional script \textit{test.py} to load the model and do the test.

In conclusion, we worked together as a team for this reproducibility companion paper. With dedication from both the original authors and reviewers, we finally obtained the improved paper and materials. We hope this paper would be very helpful for other researchers to utilize this paper.

\section{Conclusion}
In this paper, we document the replication of the original KERN method for fashion trend forecasting proposed in ICMR 2020. The replication shows that the experimental results are not strictly identical but quite similar with those reported in the original paper. Therefore, it supports the main claims in the original paper that KERN method outperforms all compared methods and the technical contributions are demonstrated to be valid.

\section*{acknowledgement}
This research is supported by the National Research Foundation, Singapore under its International Research Centres in Singapore Funding Initiative. Any opinions, findings and conclusions or recommendations expressed in this material are those of the author(s) and do not reflect the views of National Research Foundation, Singapore. We also appreciate the fashion recognition API service provided by Visenze.

\bibliographystyle{ACM-Reference-Format}
\bibliography{main}

\end{document}